\documentclass[final]{cvpr}

\usepackage{times}
\usepackage{graphicx}
\usepackage{amsmath}
\usepackage{amssymb}
\usepackage{color}
\usepackage{graphics}
\usepackage{multirow}
\usepackage{subfig}
\usepackage{array}
\usepackage{colortbl}
\usepackage{cite}
\usepackage{diagbox}
\usepackage{rotating}
\usepackage{booktabs}
\usepackage{overpic}
\usepackage{contour}
\usepackage{bbding}
\usepackage{arydshln}
\usepackage{amsfonts}
\usepackage{lipsum}
\usepackage[misc]{ifsym}

\usepackage[marginal]{footmisc}

\definecolor{mygray}{gray}{.92}

\newcommand{\myPara}[1]{\vspace{.05in}\noindent\textbf{#1~}}

\newcommand\blfootnote[1]{%
\begingroup
\renewcommand\thefootnote{}\footnote{#1}%
\addtocounter{footnote}{-1}%
\endgroup
}

\makeatletter

\newcommand{\Rmnum}[1]{\expandafter\@slowromancap\romannumeral #1@}
\makeatother

\def\ourmodel{Kaleido-BERT}
\def\afak{\emph{AFAK}}

\graphicspath{{./IMG/}}
\DeclareGraphicsExtensions{.jpg,.pdf,.png}

\usepackage[pagebackref=false,breaklinks=true,colorlinks,bookmarks=false]{hyperref}

\def\cvprPaperID{1104} 
\def\confYear{CVPR 2021}

\begin{document}

\title{\ourmodel: Vision-Language Pre-training on Fashion Domain}

\author{Mingchen Zhuge$^{1,\dagger}$~
Dehong Gao$^{1,\dagger}$~
Deng-Ping Fan$^{2,\,\textcolor{magenta}{\textrm{\Letter}}}$\\ 
Linbo Jin$^1$~
Ben Chen$^{1}$~
Haoming Zhou$^{1}$~
Minghui Qiu$^{1}$~
Ling Shao$^{2~}$\\
  $^1$ Alibaba Group \quad
  $^2$ Inception Institute of AI (IIAI) \quad \\
  {\small \url{http://dpfan.net/Kaleido-BERT}}
}

\twocolumn[{%
  \renewcommand\twocolumn[1][t!]{#1}
  \maketitle
  \vspace*{-20pt}
  \begin{center}
    \begin{center}
    \begin{overpic}[width=\linewidth]{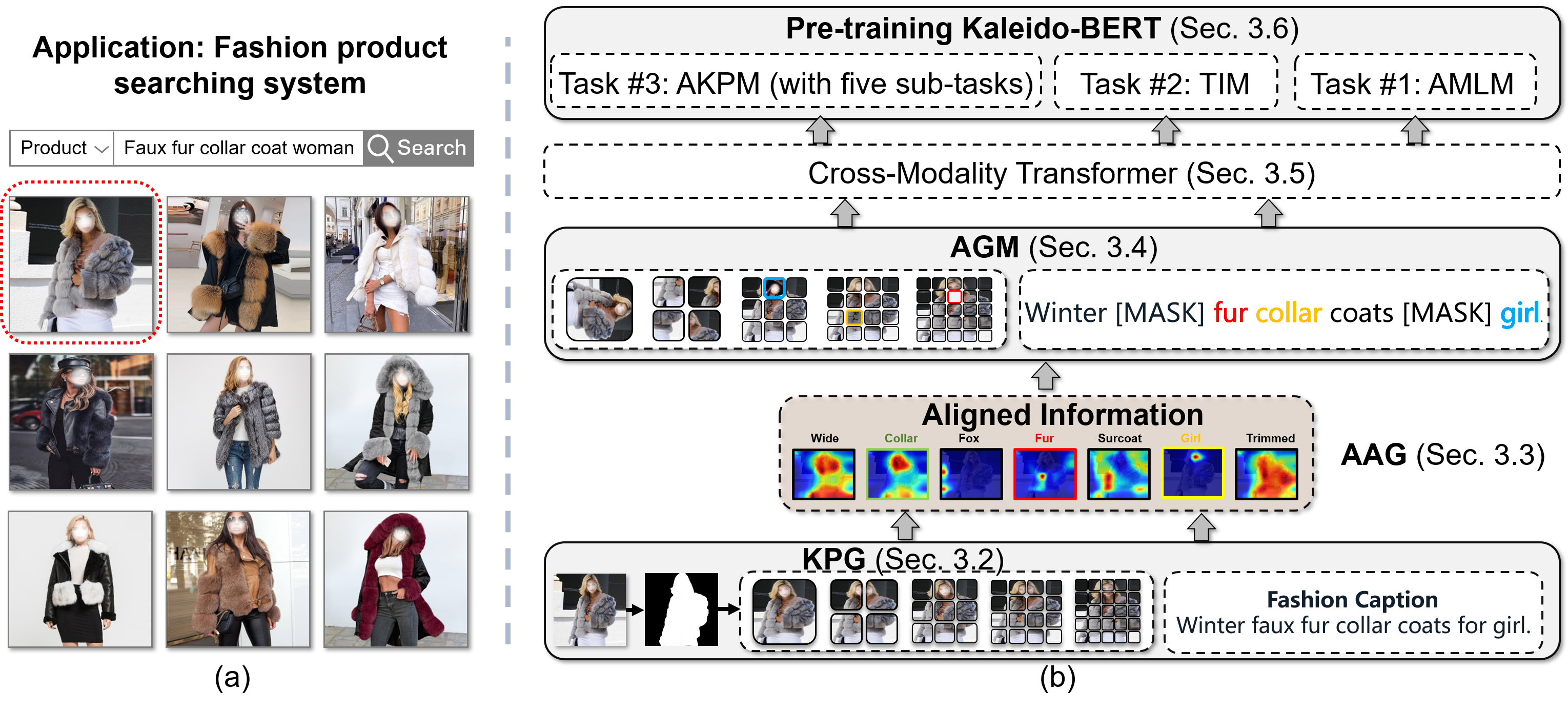}
    \end{overpic}
    \end{center}
   \vspace{-15pt}
   \captionof{figure}{\textbf{Vision-Language (VL) pre-training architecture on fashion.} We propose a novel VL pre-training architecture (\textbf{\ourmodel}), which consists of a Kaleido Patch Generator (KPG), Attention-based Alignment Generator (AAG), and Alignment Guided Masking (AGM) strategy to learn better VL feature embeddings. \ourmodel~achieves the state-of-the-art on the standard public Fashion-Gen dataset and deploys to the online system (a). 
    }
    \label{fig:Kaleido-BERT}
  \end{center}
  }]

\maketitle
\pagestyle{empty}
\thispagestyle{empty}

\begin{abstract}
We present a new vision-language (VL) pre-training model dubbed \textbf{\ourmodel}~\blfootnote{$\dagger$ Equal; * Corresponding author: Deng-Ping Fan (dengpfan@gmail.com).}, which introduces a novel kaleido strategy for fashion cross-modality representations from transformers. 
In contrast to random masking strategy of recent VL models, 
we design alignment guided masking to jointly focus more on image-text semantic relations. 
To this end, we carry out five novel tasks, \ie, rotation, jigsaw, camouflage, grey-to-color, and blank-to-color for self-supervised VL pre-training at patches of different scale.
\ourmodel~is conceptually simple and easy to extend to the existing BERT framework, it attains state-of-the-art results by large margins on four downstream tasks, including text retrieval (R@1: 4.03\% absolute improvement), image retrieval (R@1: 7.13\% abs imv.), category recognition (ACC: 3.28\% abs imv.), and fashion captioning (Bleu4: 1.2 abs imv.).
We validate the efficiency of \ourmodel~on a wide range of e-commercial websites, demonstrating its broader potential in real-world applications.
%
\end{abstract}

 \vspace{-10pt}
\section{Introduction}

%

%
Transformers~\cite{devlin2019bert,vaswani2017attention}, first designed for Natural Language Processing (NLP), have achieved great success in a number of other areas as well~\cite{brown2020language,Krzysztof2020Rethinking}, including the vision (\eg, Selfie~\cite{trinh2019selfie}, DETR~\cite{carion2020end}, ViT~\cite{vit2021ICLR}, and PVT~\cite{wang2021pyramid}) and vision-language (ViLBERT~\cite{lu2019vilbert}, VL-BERT~\cite{su2020vl}, OSCAR~\cite{li2020oscar}) communities.
However, for VL Pre-Training Model (PTM), current approaches, such as VL-BERT~\cite{su2020vl} and UNITER~\cite{chen2020uniter} focus on learning text and image representation of a general domain (\ie, coarse matching). As such, these techniques will benefit for general cross-modality representation learning. 

\begin{table*}[t]
  \centering
  \scriptsize
  \resizebox{0.99\textwidth}{!}{
    \setlength\tabcolsep{3pt}
    \renewcommand\arraystretch{1.0}
  \begin{tabular}{cr||clcccccccc}
  \toprule
   \!\!No.\!\! & Pre-Training Model & Year & Pub. & Architecture &Training Set &Core Idea&\!\!\!\!Domain\!\! &\!Pre-train\! \! &Finetune & Vision Feature &Code\\
  \hline
  \hline
  1& VisualBERT~\cite{li2019visualbert}    & 2019 & arXiv & 1-Stream & Coco & First Image-Text PTM \afak & Image &  \checkmark  & U & RoI & Torch\!\!\\
  \rowcolor{mygray}
  2& CBT~\cite{sun2019learning}  & 2019 & arXiv & 2-Stream & Kinetics~\cite{kay2017kinetics} & Noise contrastive estimation loss           & Video &  \checkmark        &    U/G/O              & Frame& N/A\!\!\\
  3& VideoBERT~\cite{sun2019videobert}       & 2019 & ICCV & 1-Stream &  SC & First Video-Text PTM \afak & Video &\checkmark  &  G/O               & Frame  & N/A\!\! \\
  \rowcolor{mygray}
  4& B2T2~\cite{alberti2019fusion}      & 2019 &EMNLP& 1-Stream &  CC&   Explicitly reference RoI to text before inputting PTM        & Image           &\checkmark& U & RoI  &Tensorflow\!\!\\
  5& LXMERT~\cite{li2020unicoder}           & 2019 & EMNLP & 2-Stream  & VG+Coco & Three encoders for ROIs, language, and cross-modality features           & Image &\checkmark   & U & RoI &Torch\!\!\\
  \rowcolor{mygray}
  6& ViLBERT~\cite{lu2019vilbert}   & 2019 & NeurlIPS & 2-Stream &  CC &  Cross-modality co-attention layers      & Image &\checkmark    &  U               &  RoI & Torch\!\! \\
  7& ImageBERT~\cite{qi2020imagebert}           & 2020 &  arXiv &  1-Stream & CC+VG+SC & Pre-training with a large-scale Image-Text dataset & Image  &\checkmark& U &  RoI     & N/A\!\!\\
  \rowcolor{mygray}
  8& Unicoder-VL~\cite{li2020unicoder}          & 2020 & AAAI& 1-Stream & CC+SBU & Masked object classification      & Image & \checkmark    &   U   & RoI  &N/A\!\! \\
  9& VLP~\cite{zhou2020unified}       & 2020 & AAAI  & 1-Stream & CC & Unified en/decoder PTM for VL generation and understanding  & Image &\checkmark         &   U/G       & RoI  &Torch\!\!\\
  \rowcolor{mygray}
  10&VL-BERT~\cite{su2019vl}       & 2020 & ICLR & 1-Stream   &CC  &  Visual feature embedding & Image &\checkmark& U & RoI &Torch\!\!\\
  11& VD-BERT~\cite{wang2020vd} & 2020 & EMNLP  & 1-Stream & VisDial~\cite{das2017visual} & Video-Dialog pre-training  & Video &  \checkmark & O & RoI &Torch\!\!\\
  \rowcolor{mygray}
  12& VLN-BERT~\cite{majumdar2020improving}    & 2020 &  ECCV & 2-Stream & Matterport3D~\cite{chang2017matterport3d} & Path selection in VL navigation       & Image &  & O &  RoI & N/A\!\!\\
  13& HERO~\cite{li2020hero}      & 2020 & EMNLP & 3-Stream & TV+HT100M & Video-subtitle matching \& Frame order modeling & Video &  \checkmark     & U & Frame  &N/A\!\!\\
  \rowcolor{mygray}
  14& XGPT~\cite{xia2020xgpt}            & 2020 & arXiv & 1-Stream & CC+SC & Improve VL generative ability   & Image & \checkmark  & U/G & RoI &N/A\!\!\\
  15& InterBERT~\cite{lin2020interbert}       & 2020 & arXiv & 1-Stream & Coco+CC+SBU & Masked group modeling  & Image &\checkmark& U & RoI  &Torch\!\!\\
  \rowcolor{mygray}
  16& VILLA~\cite{gan2020large}         & 2020 & NeurlIPS  & 2-Stream & Coco+CC+SBU & Adversarial pre-training and finetune    & Image &\checkmark& U/O & RoI &Torch\!\!\\
  17& ActBERT~\cite{zhu2020actbert}         & 2020 & CVPR  & 1-Stream &HT100M  & Global frame and local object regions \& Tangled transformer     & Video & \checkmark& U/O & Frame \& RoI &N/A\!\!\\
  \rowcolor{mygray}
  18& PREVALENT~\cite{hao2020towards}  & 2020 & CVPR & 2-Stream & Matterport3D~\cite{chang2017matterport3d} & Pre-train with image-text-action triplets       & Image & \checkmark & O  & Image &Caffe \& C++\\
  19& 12-IN-1~\cite{lu202012}            & 2020 & CVPR & 2-Stream& ES  & Multi-task training & Image & \checkmark &  U & RoI & Torch\!\!\\
  \rowcolor{mygray}
  20& Pixel-BERT~\cite{huang2020pixel}          & 2020 & arXiv& 1-Stream & Coco+VG   & Pixel-level VL semantics alignment & Image &\checkmark &  U  &  Pixel & N/A\!\!\\
  21& FashionBERT~\cite{gao2020fashionbert}      & 2020 & SIGIR & 1-Stream & Fashion-Gen~\cite{rostamzadeh2018fashion} & Patches \& Adaptive loss    & Image &\checkmark & U & Patch  & Tensorflow\\
\rowcolor{mygray}
  22& UNITER~\cite{chen2020uniter}& 2020& ECCV  & 1-Stream & Coco+VG+CC+SBU & Conditional masking \& Word region alignment & Image & \checkmark & U  &RoI  &Torch\!\!\\
  23& VisDial-BERT~\cite{murahari2019large}       & 2020 & ECCV & 2-Stream &CC+VQA~\cite{antol2015vqa}  & Adapt ViLBERT for visual dialog     & Image &    &    O              & RoI    &Torch\!\! \\
  \rowcolor{mygray}
  24& OSCAR~\cite{li2020oscar}            & 2020& ECCV  &1-Stream  & ES & Object tags as anchor points & Image &\checkmark&     U/G           & RoI & Torch\!\!\\
  25& ERNIEL-VIL~\cite{yu2020ernie}     & 2020 & arXiv & 2-Stream & CC+SBU & VL PTM with knowledge-enhanced ERNIE~\cite{zhang2019ernie} & Image & \checkmark  & U & RoI  & Paddle\!\! \\
  \rowcolor{mygray}
  26& RVL-BERT~\cite{chiou2020rvl}            & 2020 & arXiv & 1-Stream & VDR~\cite{lu2016visual} & Visual relationship detection with VL-BERT & Image  &  &       U          & RoI & Torch\!\!\\
  27& UniVL~\cite{luo2020univilm}& 2020 & arXiv & 2-Stream & HT100M & Five pre-training objectives \& Two pre-training strategies & Video & \checkmark   &          U/G        &   Frame    & N/A \\
  \rowcolor{mygray}
  28& MMFT-BERT~\cite{khan2020mmft}          & 2020 & EMNLP & 3-Stream & TV & Multi-modal fusion PTM  & Image &\checkmark& U  & RoI & Torch\!\!\\

  \midrule
 29& \textbf{\ourmodel~(OUR)}           & 2021 & CVPR & 1-Stream & Fashion-Gen~\cite{rostamzadeh2018fashion} & Kaleido patches \& Pre-alignment with masking & Image & \checkmark & U/G  & Patch \& Coordinate & Tensorflow \\
  \bottomrule
  \end{tabular}
  }
  \vspace{-10pt}
  \caption{\small \textbf{Summary of 28 previous representative cross-modality methods and our \ourmodel~model.}
  \textbf{Training Set:} Coco = \textit{MSCOCO Caption}~\cite{chen2015microsoft}. VG = \textit{Visual Genome}~\cite{krishna2017visual}. CC = \textit{Conceptual Caption}~\cite{sharma2018conceptual}. SBU = \textit{SBU Captions}~\cite{ordonez2011im2text}. TV = \textit{TVQA}~\cite{lei2018tvqa}. HT100M = \textit{HowTo100M}~\cite{miech2019howto100m}. SC: \textit{Self Collection}. ES: \textit{12-in-1 and OSCAR ensemble 12, 5+ datasets, respectively.} \textbf{Finetune:} U = Understanding tasks (\eg classification). G = Generation tasks (\eg image caption).  O = Others (\eg action task). }\label{tab:ModelSummary} 
  \vspace{-10pt}
\end{table*}

However, in the various e-commercial situations (\eg, accessories, clothing, toys), the main goal is to learn the \textit{fine-grained} representation (\eg short sleeve, cotton and jersey) rather than only the \textit{coarse} representation (what, where) in the general domain. 
In this case, the current general VL models~\cite{chen2020uniter,su2020vl} are sub-optimal for fashion-based tasks~\cite{hsiao2019fashion++,al2020paris,vasileva2018learning}, and could be unfavorable when deploying global features based models to attribute-aware tasks, such as searching for a specific fashion captioning~\cite{yang2020fashion} and fashion catalog/object~\cite{dodds2020modality}, where it is essential to extract fine-grained features or similarities~\cite{tan2019learning} from image and text.

In this paper, we propose a novel framework (see \figref{fig:Kaleido-BERT}) for the fashion-based tasks. The core idea is to focus on fine-grained representation learning and to bridge the semantic gaps between text and image. 
To achieve this goal, we first introduce an efficient ``\textit{kaleido}'' strategy, which extracts a series of multi-grained image patches for the image-modality. As a result, our model is named as \textbf{\ourmodel}.
This strategy is scalable and largely alleviates the aforementioned coarse presentation issue by introducing the patch-variant pre-training scheme.
Furthermore, to bridge the semantic gap between different modalities, attention mechanism is employed to build pre-alignments between kaleido patches and text tokens. This pre-alignment information further guides the masking strategy for pre-training. \ourmodel\footnote{\url{https://github.com/mczhuge/Kaleido-BERT/}.}~is forced to \textit{explicitly} learn semantic information across modalities.
In summary, our contributions are as follows:

\begin{figure*}[t]
	\centering
	\includegraphics[width=\textwidth]{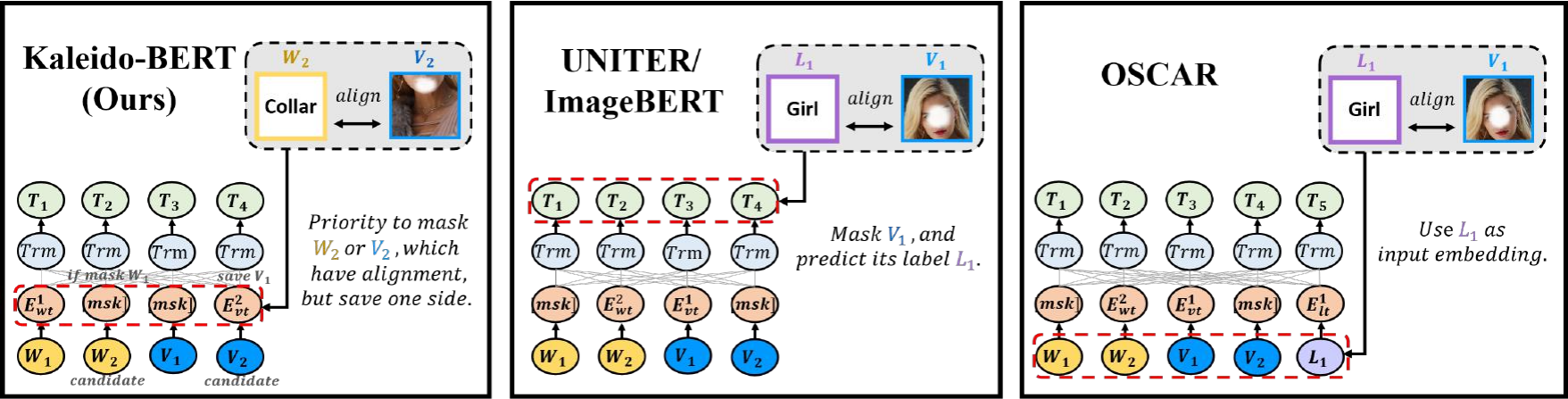}
	\vspace{-15pt}
	\caption{\small \textbf{Different utilization of alignment information in VL pre-training architectures.} \textbf{T} = \textit{Task}. \textbf{W} = \textit{Word}. \textbf{V} = \textit{Visual Token}. \textbf{L} = \textit{Object Detection Label}. \textbf{E} = \textit{Embedding}. \textbf{Trm} = \textit{Transformer Block}. 
	}
	\label{fig:Alignment}
	\vspace{-5pt}
\end{figure*}

\begin{itemize}
\vspace{-5pt}
\item \textbf{Kaleido Patch Generator}: We propose the kaleido strategy to generate a kaleido of multi-grained features. With the related pre-training tasks, \ie rotation, jigsaw, camouflage,  grey-to-color, and blank-to-color, \ourmodel~learns fine-grained cross-modality information and outperforms the fixed-patch or RoI-based VL models in the fashion domain.
\vspace{-5pt}
\item \textbf{Attention-based Alignment Generator}: \ourmodel~introduces the pre-alignment strategy to infer a cross-modality mapping between kaleido patches and text tokens. These pre-alignment pairs largely fill the semantic gaps between modalities.
\vspace{-5pt}
\item \textbf{Alignment Guided Masking}: 
We present an alignment-guided masking strategy to explicitly force \ourmodel~to learn the semantic connections between vision and language. 
Experiments show the importance of the attention-based pre-alignment and the alignment masking strategy.
\vspace{-5pt}
\end{itemize}

\section{Related Work}
There is a large body of VL modeling literature~\cite{kazemzadeh2014referitgame,antol2015vqa,plummer2015flickr30k,yang2016stacked,goyal2017making,anderson2018bottom,jiang2018pythia,suhr2018corpus,zellers2019recognition},
and we briefly introduce the transformer-based methods in this section. More detailed summary can be found in \tabref{tab:ModelSummary}.

\subsection{Vision-Language Pre-training}\label{sec:VL pretraining}
Recent transformer-based pre-training frameworks, such as  BERT~\cite{devlin2019bert}, GPT2~\cite{radford2019language}, XLNet~\cite{yang2019xlnet}, and GPT3~\cite{brown2020language}, have revolutionized NLP tasks. Motivated by these studies, many cross-modal pre-training models for vision-language (\eg, video/image and text pairs) have been designed. For video-text pair models, CBT~\cite{sun2019learning} and VideoBERT~\cite{sun2019videobert} are pioneering work that study the capability of pre-training learning.  ActBERT~\cite{zhu2020actbert} and HERO~\cite{li2020hero} focus more on downstream applications, while UniVL~\cite{luo2020univilm} focuses on both video-language understanding and generation tasks. 

For image-text pair models, they can be categorized into single-stream~\cite{li2019visualbert,alberti2019fusion,li2020unicoder,zhou2020unified,su2019vl,wang2020vd,xia2020xgpt,lin2020interbert,huang2020pixel,li2020does,chen2020uniter,gao2020fashionbert,li2020oscar,chiou2020rvl,qi2020imagebert} and two-stream~\cite{li2020unicoder,lu2019vilbert,murahari2019large,majumdar2020improving,desai2020virtex,hao2020towards,dodds2020modality,yu2020ernie} or even three-stream~\cite{khan2020mmft} according to the  network architecture of the single-modal input.
In single-stream models, the features of different modalities are directly fed into a Transformer. 
In contrast, in two-stream models, they are first processed by two single-modal networks before fed into a Transformer, and so forth in three-stream models. 
ViLBERT~\cite{lu2019vilbert} claims that the two-stream structure is superior to the single-stream, 
while VL-BERT~\cite{su2019vl} finds that the single-stream models achieve more promising results, as these models have more cross-modality information interactions. 
VisualBERT~\cite{li2019visualbert} and B2T2~\cite{alberti2019fusion} are single-stream models and derive a unified VL understanding network. With the same spirit of focusing on generic VL tasks, many concurrent models, \eg, Unicoder-VL~\cite{li2020unicoder}, VLP~\cite{zhou2020unified}, ViLBERT~\cite{lu2019vilbert},  VL-BERT~\cite{su2019vl}, have been proposed under the BERT framework. 
In contrast to the boom in generic tasks (\eg, VCR~\cite{chen2020uniter,yu2020ernie,lin2020interbert}, VQA~\cite{khan2020mmft,li2020unicoder}), other tasks such as visual relationship detection (RVL-BERT~\cite{chiou2020rvl}), visual navigation (\ie, PERVALENT~\cite{hao2020towards} and VLN-BERT~\cite{majumdar2020improving}), and visual dialog (\eg, VisualD~\cite{murahari2019large}, VD-BERT~\cite{wang2020vd}) are still in their infancy.
More recently, Lu \etal~\cite{lu202012} shows that multi-task VL learning can lead to a significant improvement over isolated task learning. Similarly, OSCAR~\cite{li2020oscar} achieves new state-of-the-art performance on many representative VL tasks (\eg, image captioning like XGPT~\cite{xia2020xgpt}, image-text retrieval like Image-BERT~\cite{qi2020imagebert}).

Advances have also been made by the VirTex~\cite{desai2020virtex} model in image classification, object detection, and instance segmentation filed, by using semantically dense captions to learn visual representations. Another notable study is the recent ACL work~\cite{li2020does} in which the authors creatively demonstrate that the attention head of the VL model can perform entity grounding and syntactic grounding. Unlike all the above-mentioned works, Pixel-BERT~\cite{huang2020pixel} considers aligning vision-language features at a pixel level instead of using region-based image features.

As shown in \figref{fig:Alignment}, our \ourmodel~focuses on a masking strategy at the embedding level rather than at the task level (\eg, LXMERT~\cite{li2020unicoder} and UNITER~\cite{chen2020uniter}) or input level such as OSCAR~\cite{li2020oscar}. \ourmodel~\textit{explicitly} align the embedding features between image and text so that it can learn fine-grain representations for fashion tasks.

\subsection{Fashion-Based Task}
As described in \secref{sec:VL pretraining}, existing VL models mainly focus on relatively \textit{coarse representations}, while less attention has been paid to fine-grained representation learning for the fashion-based task. There are two concurrent studies~\cite{dodds2020modality,gao2020fashionbert} resembling our work. FashionBERT~\cite{gao2020fashionbert} was the first published work in the fashion domain.  
The concurrent work, MAAF~\cite{dodds2020modality}, 
aims to derive a modality-agnostic attention fusion strategy to address the undifferentiated text and image query task. Unlike FashionBERT, which utilizes a patch-fixed masking strategy, the MAAF adopts an image-level attention mechanism. We argue that these two schemes restrict the power of the pre-trained representation learning, especially for the fine-grained fashion task. As a consequence, a more flexible solution with patch-variant is urgently required.

To the best of our knowledge, the proposed \ourmodel~is the first to present the effectiveness of alignment guided masking by jointly focusing more on image-text coherence for the fashion domain.

\section{Proposed \ourmodel}
In this section, we introduce our \ourmodel, which learns the \textit{fine-grained} VL features for the fashion domain rather than the \textit{coarse representation} features for VL tasks. We adopt the standard transformer designed for NLP to make our \ourmodel~scalable over a varying number of transformer-based VL learning task.

\subsection{Model Overview}
The architecture of our \ourmodel~is illustrated in \figref{fig:Kaleido-BERT}. There are five stages: 
(1) \ourmodel~takes two inputs: a text (\eg, image caption or description) and corresponding image patches generated by our Kaleido Patches Generator (\textbf{KPG}). Similar to LXMERT~\cite{li2020unicoder}, each text is represented as a sequence of tokens and each image is represented as a sequence of kaleido patches. 
(2) At the embedding stage, we propose the Attention-based Alignment Generator (\textbf{AAG}) to generate pre-alignments between text tokens and kaleido patches so that the image and text are explicitly aligned semantically.
(3) Different from existing random masking strategy, we proposed 
to adopt an Alignment Guided Masking (\textbf{AGM}) strategy to relieve the difficulty of cross-modality modeling.
(4) Text tokens and kaleido patches fully interact in \ourmodel, which gradually learns VL semantic information and produces the cross-modality fine-grained representations.
(5) Finally, we adopt five new kaleido  tasks (\ie, rotation, jigsaw, camouflaged, grey-to-color and blank-to-color tasks) besides the masked language modeling and image-text matching tasks to supervise the network. 
Our implementation is based on the \texttt{EasyTransfer~\footnote{Tensorflow: \url{https://github.com/alibaba/EasyTransfer}}/Huggingface~\footnote{ Pytorch: \url{https://github.com/huggingface/transformers}}} library. We refer the readers to 
this de facto standard library for details. 

\subsection{Kaleido Patch Generator}\label{sec:KPG}
\begin{figure}[t!]
	\centering
	\includegraphics[width=\columnwidth]{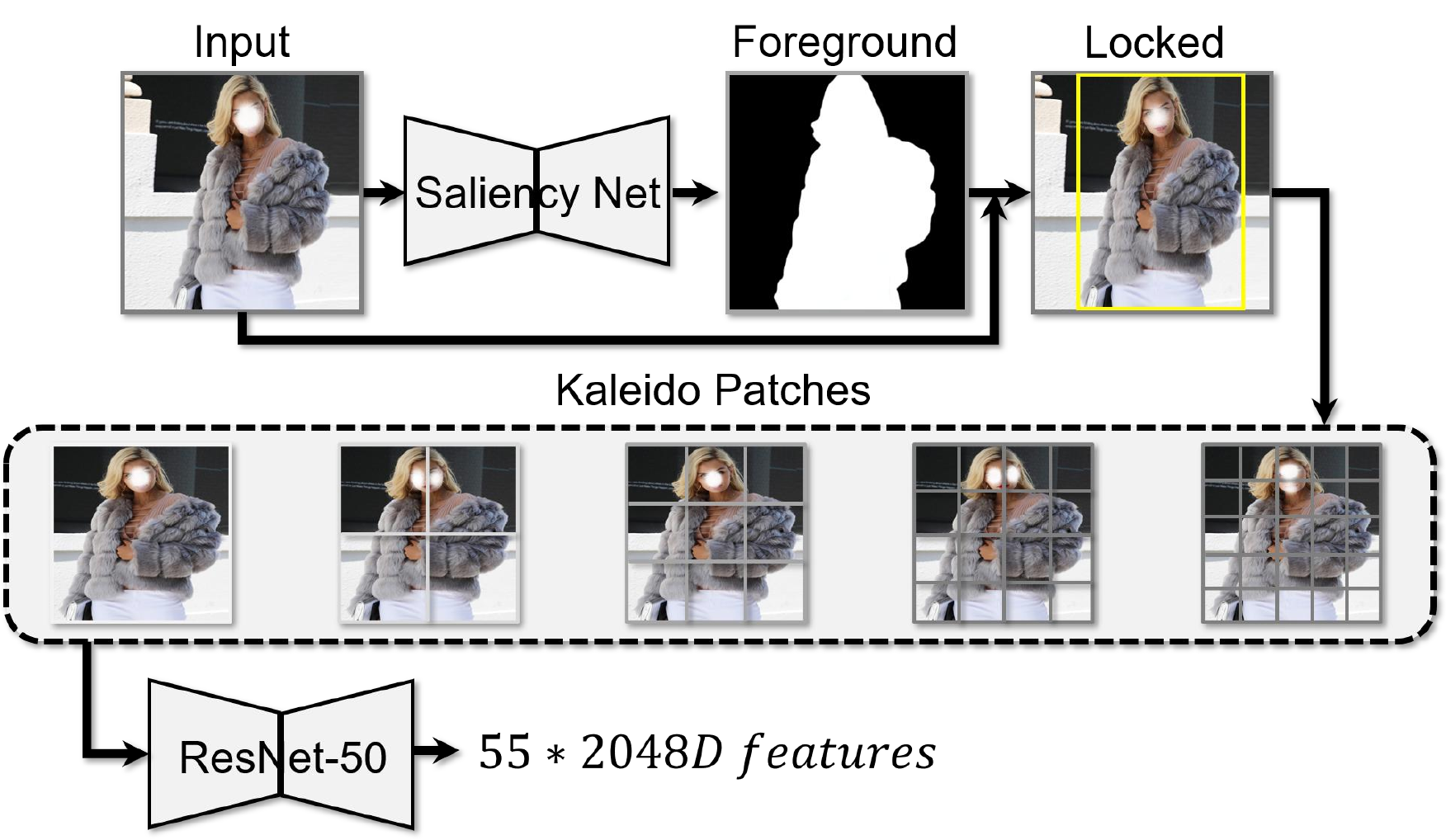}
	\vspace{-20pt}
	\caption{\small \textbf{Illustration of KPG.} See \secref{sec:KPG} for details.}
	\label{fig:KPG}
\end{figure}

Given an image as input, we obtain the multi-grained patches by the proposed Kaleido Patch Generator (\textbf{KPG}). As shown in \figref{fig:KPG}, we can introduce a saliency detection network\footnote{For simplicity, we just utilize a very simple UNet-like architecture as our foreground segmentation net.} (\eg, BAS~\cite{qin2021boundary}, EGNet~\cite{zhao2019egnet}, ICON~\cite{zhuge2021salient} or other SOTAs in the recent paper~\cite{fan2018salient}) to obtain the foreground mask and then lock (\eg, bounding box proposal) the domain object. Motivated by the spatial envelope~\cite{oliva2001modeling} and the block-level strategy~\cite{fan2019scoot,fan2017structure}, we then split the image into different scales (\ie, 1$\times$1, 2$\times$2, $\dots$, 5$\times$5). These image patches are just like ``\textit{kaleido}'' patches, and 
more detailed divisions (\eg, 6$\times$6 or N$\times$N like Pixel-BERT~\cite{huang2020pixel}) can be considered according to the difficulty of the specific task. Finally, we obtain 55 kaleido patches from each input image. To generate the embeddings of these patches, we utilize the standard ResNet-50~\cite{he2016deep} as our backbone.

\begin{figure}[t!]
	\centering
	\includegraphics[width=\columnwidth]{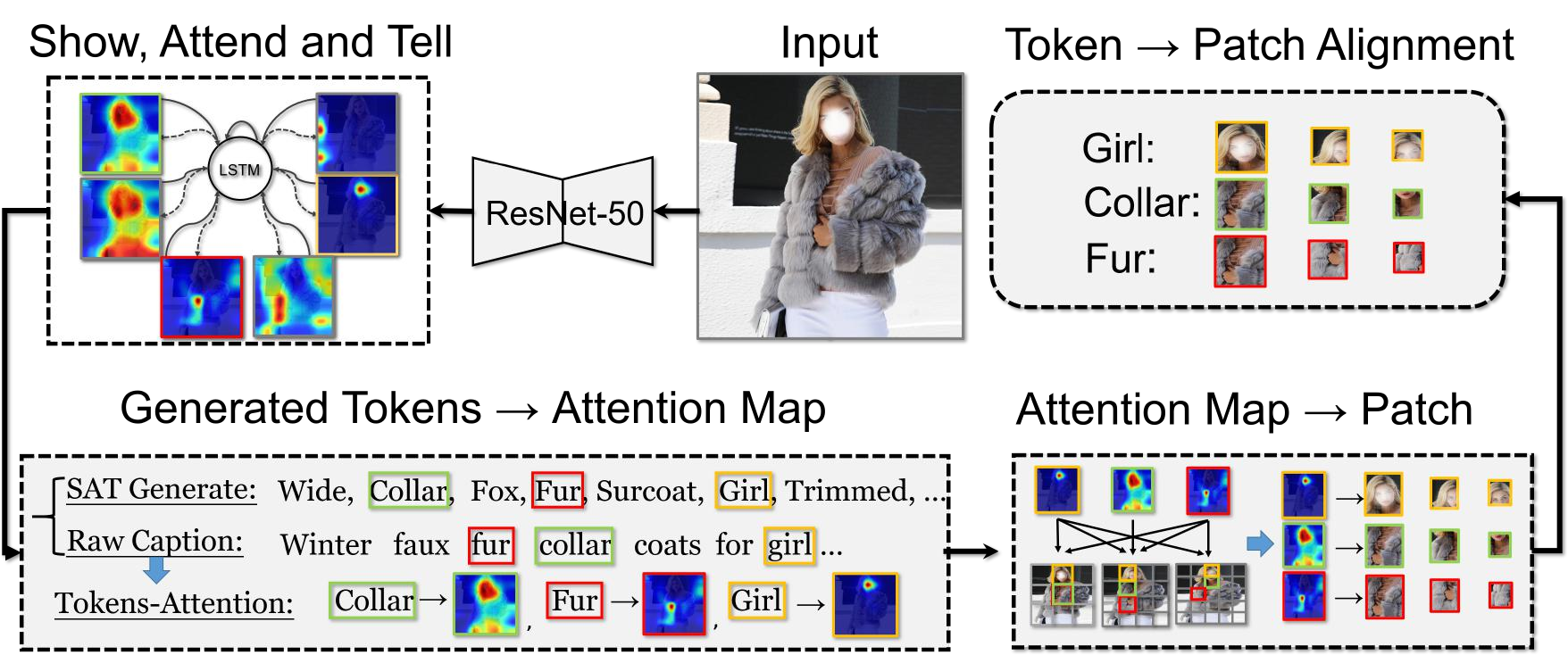} 
	\caption{\small \textbf{Procedures of AAG.} See \secref{sec:AAG} for details.
	}\label{fig:AAG}
\end{figure}

\subsection{Attention-based Alignment Generator}\label{sec:AAG}

Attention-based Alignment Generator (\textbf{AAG}) aims to find the coarse alignments between text tokens and kaleido patches.
As shown in \figref{fig:AAG}, we directly adopt the famous SAT network~\cite{xu2015show} as our text generator, which automatically learns to describe the content of images.
At the same time, the SAT network generates the attention heat-map for each token,
from which we infer the relation between generated tokens and image regions.
With the co-occurrence of the generated tokens and the raw text tokens, as well as the overlap area of image regions and kaleido patches, we further build the alignments between raw text tokens and kaleido patches.

\begin{figure*}[t!]
	\centering
	\includegraphics[width=\textwidth]{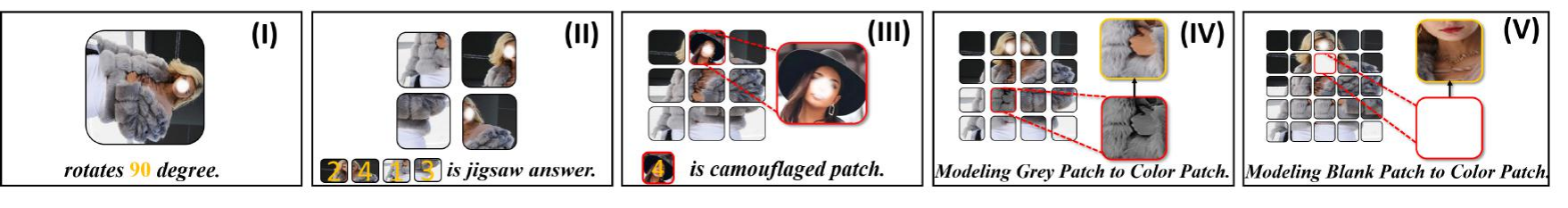}
	\vspace{-20pt}
	\caption{\small \textbf{Aligned Kaleido Patch Modeling (AKPM).} (I) Rotation recognition. (II) Jigsaw puzzles solving. (III) Camouflaged prediction. (IV) Grey-to-color modeling. (V) Blank-to-color modeling. Zoomed-in for a better view. See \secref{sec:Pre-trainKBERT} for details.}	\label{fig:kaleido}
	\vspace{-10pt}
\end{figure*}

\subsection{Alignment Guided Masking}
The main idea that inspires us to modify the vanilla random masking strategy is that the pre-aligned
$\langle$token, patch$\rangle$ pair provides explicit semantic relations between two modalities. This alignment can be 
used in the pre-training stage, which further forces \ourmodel~to \textit{explicitly} explore cross-modality semantic information.
As shown in \figref{fig:Alignment} (Left), unlike the random masking strategy, Alignment Guided Masking (\textbf{AGM}) gives high priority to masking the pre-alignment pairs. Meanwhile, for each selected pre-aligned $\langle$token, patch$\rangle$ pair, we randomly mask either the token part or the patch part, which stimulates~\ourmodel~to learn the missing information in one modality by providing the information of the other. When all pre-alignment pairs are traversed and not-enough tokens or patches are selected, a random masking strategy is adopted to mask the unaligned tokens and patches independently. In this way, we obtain the token and patch masking candidates. AGM strategy works on level-3, level-4, level-5 of kaleido patches.
We do not apply this strategy on level-1 \& -2 since masking larger patches will increase the difficulty of modeling. Empirically, we mask one patch in level-3, two patches in level-4, and three patches in level-5.
%

\subsection{Cross-Modality Transformer}
We adopt the original BERT~\cite{devlin2019bert} as our cross-modality transformer so that our \ourmodel~can be easily extended. Specifically, for the text side, we follow FashionBERT~\cite{gao2020fashionbert} to encode the order of the token (\ie, generated via WordPieces~\cite{wu2016google}) position as $0,1,2,3, \dots,N$. 
Our final training corpus for each sub-word token is obtained by summing up its embedding with the segment and position embeddings, followed by another layer normalization (LN) layer. For the image side, we encode the position information by re-organizing it as 5D features ($[x_1,x_2,y_1,y_2,w*h]$) for each patch.
After that, both patches and location features are fed into a fully-connected (FC) layer in order to project them into the same embedding space. 
We obtain visual embeddings for each patch by summing up three 
FC outputs (\ie, FC (seg\_id), FC (img\_feature), FC (pos\_emb))\footnote{Similar to `segment embeddings' in BERT, we conduct a special modality embedding (`T' for text, `I' for image) to help the model distinguish the different modalities.} and then passing them through an LN layer.

\subsection{Pre-training \ourmodel}\label{sec:Pre-trainKBERT}
To alleviate the VL semantic gap and boost feature representation, we design three pre-training tasks, \ie, Aligned Masked Language Modeling (\textbf{AMLM}), Image and Text Matching (\textbf{ITM}) and the proposed Aligned Kaleido Patch Modeling (\textbf{AKPM}) (which includes five kaleido sub-tasks) to supervise our \ourmodel. 

\myPara{Task \#1: AMLM.}
Derived from our alignment guided masking strategy, we can obtain the mask candidates including both token and patch candidates. When masking indices are determined, we decompose masking words into 10\% random words, 10\% unchanged, and 80\% \texttt{[MSK]}. 
The masked-out token sequence is denoted by $T_{i} = \left\{t_1,...[MSK],...,t_T\right\}$, where token $t_i$ is masked out.
We feed the hidden output of the last layer of the masked-out token into a classifier over the standard BERT vocabularies.
The AMLM goal is to predict the masked words based on the observation of their surrounding tokens and image patches. 
The objective of the AMLM task is mathematically written as: 
\begin{equation}
\small
\mathcal{L}_{AMLM}=\sum CE(
    t_{i}, \mathcal{F}(T,K, \theta)_{MSK\_hidden}),
\end{equation}
where $CE$ denotes the cross-entropy loss. 
$\mathcal{F}$ is the~\ourmodel~function. 
$\mathcal{F}(\cdot)_{MSK\_hidden}$ denotes the hidden output of masked-out tokens. 
$K$ denotes the masked-out kaleido patch sequence.

\myPara{Task \#2: ITM.}
The ITM task is transferred by Next Sentence Prediction (NSP) on the vanilla BERT. In this task, \texttt{[CLS]} is used to indicate the beginning of the fused representation. 
The hidden output of \texttt{[CLS]} is fed into an FC layer and we use the sigmoid function to predict a score between 0 and 1. The text and image of one positive example are extracted from the same fashion product, while those of one negative sample are randomly extracted from different fashion products. The objective of the ITM task is written as:
\begin{equation}
\small
\mathcal{L}_{ITM}=CE(y_{m}, \mathcal{F}(T,K,\theta)_{CLS\_hidden}),
\end{equation}
where $y_m$ denotes the text and image match label.

\myPara{Task \#3: AKPM.}
The kaleidoscope patch sequence is composed of a collection of kaleidoscope patches as $\{K_{1}, K_{2}, ..., K_{N}\}$, in which $N$ is the kaleidoscope level ($N=5$ in our experiment). 
As shown in \figref{fig:kaleido}, our AKPM includes a single sub-task for each kaleidoscope level, respectively.

\myPara{Sub-Task \#\Rmnum{1}: Rotation Recognition (RR).} 
Recent works ~\cite{goyal2019scaling, jing2020self} have compared various self-supervised learning strategies concluding that predicting image rotations is 
among the most effective.
Motivated by this, we introduce RR in our pre-training.
Specifically, we force the 1$\times$1 patch of the level-1 kaleido to randomly rotate by an angle $ \theta \in \left\{0^\circ, 90^\circ, 180^\circ, 270^\circ\right\}$. 
During the training process, we use the angle of the rotated patch as the target label. The hidden output of the $K_{1}$ patch is fed into an FC layer followed by softmax function. The final softmax output is used to predict the angle. 
The objective of the RR task is written as:
\begin{equation}
\small
\mathcal{L}_{RR}=CE(y_{r}, \mathcal{F}(T,K,\theta)_{K_1\_hidden}),
\end{equation}
where $y_r$ denotes the rotation angle.

\myPara{Sub-Task \#\Rmnum{2}: Jigsaw Puzzle Solving (JPS).} 
JPS~\cite{noroozi2016unsupervised, jing2020self} has been demonstrated to be suitable for self-supervised representation learning.  Such a pretext task (also called surrogate task) can mine the spatial relations among image patches. Based on this insight, we borrow the notion of jigsaw puzzle to stimulate~\ourmodel~to learn the potential association from unordered 2$\times$2 patch lists.  
For simplicity, we treat the JPS problem as a classification of the jigsaw permutations ($4!=24$ classes). The network architecture is similar to RR. The objective of the JPS task is written as:
\begin{equation}
\small
\mathcal{L}_{JPS}=CE(y_{j}, \mathcal{F}(T,K,\theta)_{K_2\_hidden}),
\end{equation}
where $y_j$ denotes the jigsaw permutation.

\myPara{Sub-Task \#\Rmnum{3}: Camouflage Prediction (CP).} 
To increase the discernment ability of the model, we introduce another camouflage prediction task to judge which patch has been replaced\footnote{The camouflaged patch is the same scale as the patch randomly selected selecting from the other product randomly.}. 
With the help of image and text clues, this task encourages the training process to observe the diversity among 3$\times$3 patches. We name this task Camouflage Prediction (CP) because its essence is to camouflage one patch then let the model detect it. By pre-training our \ourmodel~with CP, the framework achieves a strong capacity to screen out the imparity with varied products. 
The CP prediction is also treated as a classification problem and its objective is denoted by:
\begin{equation}
\mathcal{L}_{CP}=CE(y_{c}, \mathcal{F}(T,K,\theta)_{K_3\_hidden}),
\end{equation}
where $y_c$ denotes the index of a camouflaged patch.

\myPara{Sub-Task \#\Rmnum{4}: Grey-to-Color Modeling (G2CM).} 
Different from the masking strategy in existing models, which simply exchanges image embeddings with zero paddings, we introduce a smoother G2CM strategy that greys the image patches. Then we reconstruct the grey patch to a color patch by regression, supervised by KL-divergence, which better caters to self-supervised learning. 
The objective of G2CM is to minimize the G2CM loss:
\begin{equation}
    \mathcal{L}_{G2CM}=\sum KLD(k_{4i}, \mathcal{F}(T,K,\theta)_{K_4\_hidden}),
\end{equation}
where $KLD$ denotes the KL-divergence, which aims to minimize the distance of the reconstructed distribution to the target distribution and $k_{4i}$ is the masked-out patch(es) of $K_4$ kaleido patches.

\myPara{Sub-Task \#\Rmnum{5}: Blank-to-Color Modeling (B2CM).}
The last sub-task is B2CM. Similar to other pre-training methods that replace image feature embeddings with the same-dimension zeros sequence, we also adopt this kind of patch masking scheme. 
This strongly tests the learning ability of a model that captures the contextual information.
The objective of B2CM is to minimize the B2CM loss:
\begin{equation}
    \mathcal{L}_{B2CM}=\sum KLD(k_{5i}, \mathcal{F}(T,K,\theta)_{K_5\_hidden}),
\end{equation}
where $k_{5i}$ is the masked-out patch of $K_5$.
\\

All in all, we introduce the aligned kaleido patch modeling to enhance the ability of the model for spatial context structure (\ie, RR and JPS), classification (\ie, CP), and image generation (\ie, G2CM and B2CM). Finally, \ourmodel~should minimize the overall loss function as:
\begin{equation} 
\begin{split}
        & \mathcal{L}_{total} = \mathcal{L}_{AMLM} + \mathcal{L}_{ITM} + \mathcal{L}_{RR} + \mathcal{L}_{JSP} \\ 
      &+ \mathcal{L}_{CP} + \mathcal{L}_{G2CM} + \mathcal{L}_{B2CM}. \label{eq:total_loss}
\end{split}
\end{equation}

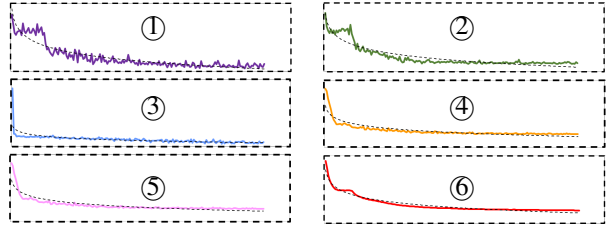
\begin{figure}[t!]
	\centering
	\begin{overpic}[width=.47\textwidth]{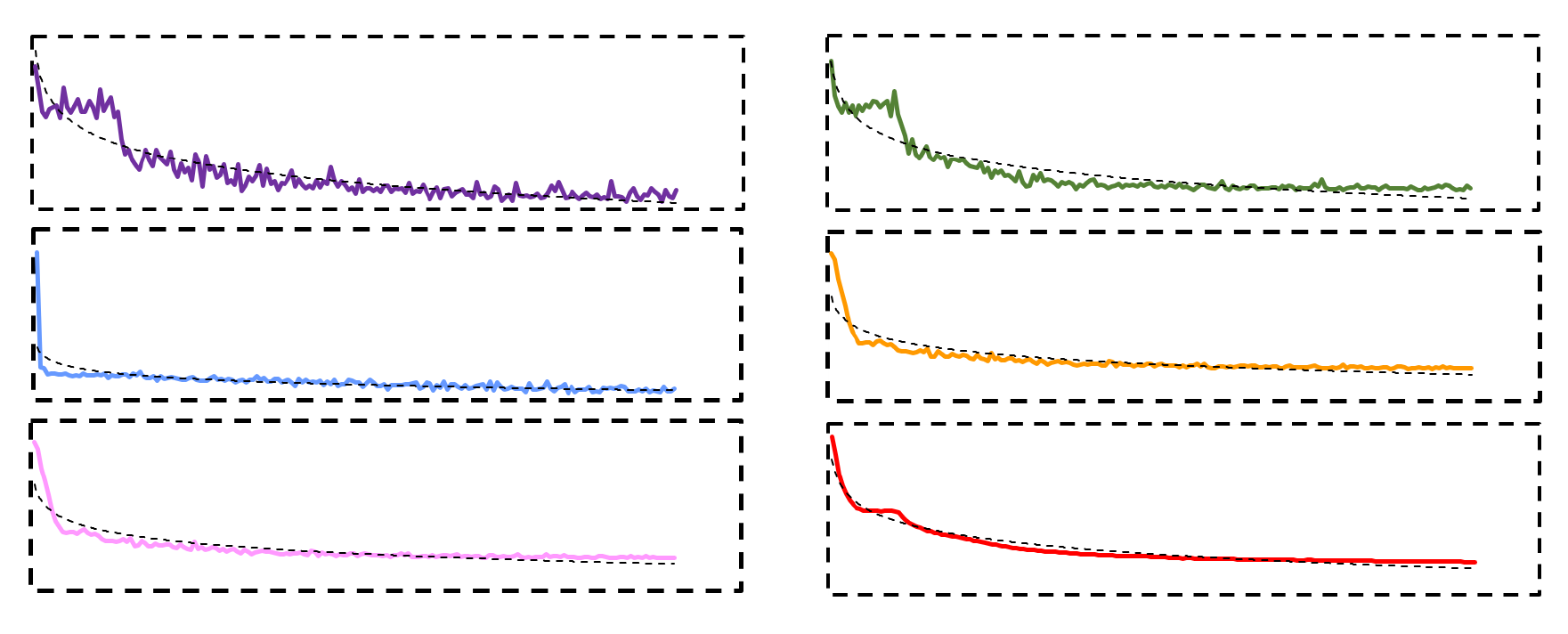}
		\put(23,33){\normalsize{\textcircled{1}}}
		\put(73,33){\normalsize{\textcircled{2}}}
		\put(23,20){\normalsize{\textcircled{3}}}
		\put(73,20){\normalsize{\textcircled{4}}}
		\put(23,7){\normalsize{\textcircled{5}}}
		\put(73,7){\normalsize{\textcircled{6}}}
	\end{overpic}
	\vspace{-10pt}
	\caption{\small \textbf{Evolution of each training loss.} The scores are computed from the validation sets. \textcircled{1}: \textit{Rotation loss.} \textcircled{2}: \textit{Jigsaw loss.} \textcircled{3}: \textit{Camouflage loss.} \textcircled{4}: \textit{Grey-to-Color loss.} \textcircled{5}: \textit{Blank-to-Color loss.} \textcircled{6}: \textit{Total Loss = AKPM + ITM + AMLM.} This shows that \ourmodel~can learn from the kaleido strategy. }
	\label{fig:loss}
	\vspace{-10pt}
\end{figure}

The evaluations of different Kaleido tasks on the validation set are shown in Fig.~\ref{fig:loss}. As can be seen, the losses decay smoothly, which proves that the pre-training process carries on as normal, and the designed tasks can be learned well with \ourmodel.

\section{Experiments}

We evaluate our \ourmodel~on four VL tasks by transferring the pre-trained model to each target task and fine-tuning through end-to-end training.

\begin{table*}[t!]
\center
\renewcommand{\tabcolsep}{2.5pt}
\renewcommand{\arraystretch}{1.0}
\caption{\small \textbf{Retrieval performances on Fashion-Gen dataset.} Here, $\text{Sum}\mathcal{R}$=$($Rank@1+Rank@5+Rank@10$)$*100. See \secref{sec:CompetingModel} for details.
}\label{tab:exp_i2t_t2i}
\vspace{-10pt}
\resizebox{1.0\textwidth}{!}
{
\begin{tabular}{llc||ccccccccc||c}
\toprule
\multicolumn{3}{c||}{} & \textbf{\textsf{\;\;\;\;VSE\;\;\;\;}} & \textbf{\textsf{\;\;\;VSE++\;\;\;}} & \textbf{\textsf{\;\;SCAN\;\;}} & \textbf{\textsf{\;\;PFAN\;\;}} & \textbf{\textsf{\;ViLBERT\;}} & \textbf{\textsf{\;VLBERT\;}} & \textbf{\textsf{FashionBERT}} & \textbf{\textsf{ImageBERT}} & \textbf{\textsf{\;OSCAR\;}} & \textbf{\textsf{\ourmodel}} \\ 
\multicolumn{3}{c||}{\multirow{-2}{*}{\textbf{\textsf{Tasks}}}} & ~\cite{kiros2014unifying} & ~\cite{faghri2017vse++} & ~\cite{lee2018stacked} & ~\cite{wang2019position} & ~\cite{su2020vl} & ~\cite{lu2019vilbert} & ~\cite{gao2020fashionbert} & ~\cite{qi2020imagebert} & ~\cite{li2020oscar} & \multicolumn{1}{c}{\textit{Ours}} \\ 
 \midrule

 & Rank@1 & $\uparrow$ & 4.010\% & 4.590\% & 4.590\% & 4.290\% & 20.97\% & 19.26\% & \underline{23.96\%} & 22.76\% & 23.39\% &~\textbf{27.99\%}{$_{(+ 4.030\%)}$}  \\
 & Rank@5 & $\uparrow$ & 11.03\% & 14.99\% & 16.50\% & 14.90\% & 40.49\% & 39.90\% & \underline{46.31\%} & 41.89\% & 44.67\% & ~\textbf{60.09\%}{$_{(+ 13.78\%)}$}  \\
\multirow{-3}{*}{\begin{tabular}[c]{@{}l@{}}\textsc{1.ITR}\\ \end{tabular}} 
 & Rank@10 & $\uparrow$ & 22.14\% & 24.10\% & 26.60\% & 24.20\% & 48.21\% & 46.05\% & 52.12\% & 50.77\% & \underline{52.55\%} &  ~~\textbf{68.37\%}{$_{(+ 15.82\%)}$ } \\
 
  \midrule
 & Rank@1 & $\uparrow$ & 4.350\% & 4.600\% & 4.300\% & 6.200\% & 21.12\% & 22.63\% & \underline{26.75\%} & 24.78\% & 25.10\% &~\textbf{33.88\%}{$_{(+ 7.130\%)}$}  \\
 & Rank@5 & $\uparrow$ & 12.76\% & 16.89\% & 13.00\% & 20.79\% & 37.23\% & 36.48\% & 46.48\% & 45.20\% & \underline{49.14\%} & ~\textbf{60.60\%}{$_{(+ 11.46\%)}$}  \\
 \multirow{-3}{*}{\begin{tabular}[c]{@{}l@{}}\textsc{2.TIR}\end{tabular}} 
 & Rank@10 & $\uparrow$ & 20.91\% & 28.99\% & 22.30\% & 31.52\% & 50.11\% & 48.52\% & 55.74\% & 55.90\% & \underline{56.68\%} & ~\textbf{68.59\%}{$_{(+11.91\%)}$}  \\  
 \midrule  
 &$\text{Sum}\mathcal{R}$& $\uparrow$ & 75.20 & 94.16 & 87.29 & 101.90 & 218.13 & 212.84 & 251.36 & 241.30 & \underline{251.53} & ~\textbf{319.52} \\
\bottomrule
\end{tabular}
}
\end{table*}

\subsection{Pre-training Settings}
\textbf{Dataset.} For a fair comparison, we follow the same settings as the Top-1 FashionBERT~\cite{gao2020fashionbert} and pre-train the proposed \ourmodel~on the Fashion-Gen\footnote{\url{https://fashion-gen.com/}} dataset. 
It contains 67,666 fashion products accompanied with text descriptions.
Each product includes one to six images from different angles. 
Among all the image-text pairs, like~\cite{gao2020fashionbert}, we use 260,480 for training, and 35,528 for testing. 

\textbf{Implementation Details.} Our \ourmodel~ has: L=12, H=768, A=12. L is number of stacked Transformer blocks. H denotes the hidden activation, and A means the number of attention heads. 
We implement our model with Tensorflow and use 8*Tesla V100 for pre-training. 
The Adam optimizer is applied with a learning rate of $2e-5$ and weight decay $1e-4$. We adopt a warming-up strategy for the first 5K steps.

\subsection{Downstream Tasks}
We evaluate our model for four downstream VL tasks, including Image-Text Retrieval, Text-Image Retrieval, Category Recognition, and Fashion Captioning. The four tasks strongly cater to industrial applications in the fashion field.

\textbf{1. Image-Text Retrieval (ITR).~}
Text retrieval is a downstream task that requires the model to distinguish whether a sentence can effectively describe an image. We sample the product images and titles as image-sentences pairs provided by the Fashion-Gen~\cite{rostamzadeh2018fashion} and consider the original product information as positive samples. At the same time, we shuffle the dataset and consider the unmatched image-sentence pairs as negative samples. To increase the difficulty, the positive and negative pairs are selected from the same sub-category, which is hard for PTM to differentiate. We use Rank@1, Rank@5, Rank@10 to evaluate the retrieval performance.

\textbf{2. Text-Image Retrieval (TIR).~}
The image retrieval task aims to rank product images according to their title. Similar to text retrieval, we use the ground-truth image in the pair as the positive sample and randomly sample 100 unrelated captions from other products in the same sub-category. By predicting the matching score, Rank@1, @5, @10 are used as metrics.

\textbf{3. Category/SubCategory Recognition (CR \& SUB).}
The category is a vital attribute for describing a product, and is especially useful in many real-life applications. 
We consider a classification task that judges the category and subcategory of a product, such as \{\texttt{HOODIES, SWEATERS}\}, \{\texttt{TROUSERS, PANTS}\}. We directly use a FC layer after \texttt{[CLS]} for these tasks.

\textbf{4. Fashion Captioning (FC).~}
Image captioning has emerged as an important research topic with a rich literature in computer vision, and the accuracy on FC can evaluate the generation ability of cross-modality models.

\subsection{Competing Models}\label{sec:CompetingModel}
Detailed comparisons for each downstream task are shown in ~\tabref{tab:exp_i2t_t2i} \& ~\tabref{tab:exp_catpred_generation}. 
($i$) Our~\ourmodel~achieves significant improvement on nearly all evaluations, which demonstrates its excellent understanding and generation ability in fashion domain.
($ii$) We observe that the FashionBERT approach outperforms ViLBERT and VLBERT. The main difference between them is that FashionBERT adopts patches as image features, while ViLBERT and VLBERT extract RoIs as features. This indicates that in the fashion domain, the patch method is better for extracting image features.
($iii$) ImageBERT and Oscar perform better than VLBERT and ViLBERT by adding RoI object classification and RoI tags. 
These two methods provide more information about the image modality. 
This, to a certain degree, hints that more image semantic information (\eg image features, image supervision tasks) is required to guide model learning.
In our~\ourmodel, the kaleido strategy extends from the patch method of FashionBERT~\cite{gao2020fashionbert}. 
The attention-based alignment masking and the kaleido pre-training task provide more semantic information from the image modality.
These factors together explain the superiority of~\ourmodel~in VL understanding and generation in the fashion domain.

\begin{table}[t!]
\center
\renewcommand{\tabcolsep}{2pt}
\renewcommand{\arraystretch}{1.0}
\caption{
\small \textbf{Category Recognition and Fashion Captioning performances on Fashion-Gen dataset.} Here, $\text{Sum~} \mathcal{C}\mathcal{L}\mathcal{S}$= $($ACC+macro-$\mathcal{F}$$)$*100 and $\text{Sum~}\mathcal{C}\mathcal{A}\mathcal{P}$=Bleu-4+METEOR+ROUGE-L+CIDEr.
See \secref{sec:CompetingModel} for details.
}\label{tab:exp_catpred_generation}
\vspace{-10pt}
\resizebox{.47\textwidth}{!}
{
\begin{tabular}{llc||ccc||l}
\toprule
\multicolumn{3}{c||}{}  & \textbf{\textsf{FashionBERT}} & \textbf{\textsf{ImageBERT}} & \textbf{\textsf{\;OSCAR\;}} & \textbf{\textsf{\ourmodel}} \\ 
\multicolumn{3}{c||}{\multirow{-2}{*}{\textbf{\textsf{Tasks}}}}  & ~\cite{gao2020fashionbert} & ~\cite{qi2020imagebert} & ~\cite{li2020oscar} & \multicolumn{1}{c}{~\textit{Ours}} \\ 
\midrule

 & ACC & $\uparrow$ &91.25\% & 90.77\% & \underline{91.79\%} & \textbf{95.07\%} {$_{(+3.28\%)}$}  \\
 \multirow{-2}{*}{\begin{tabular}[c]{@{}l@{}}\textsc{3.CR} \end{tabular}} 
 & macro-$\mathcal{F}$ & $\uparrow$  & 0.705 & 0.699 & \underline{0.727} & \textbf{0.714} ~~~~{$_{(-0.013)}$} 
\\ 
\hdashline
 & ACC & $\uparrow$  & \underline{85.27\%} & 80.11\% & 84.23\% & \textbf{88.07\%} {$_{(+2.80\%)}$}  \\
 \multirow{-2}{*}{\begin{tabular}[c]{@{}l@{}l}\textsc{3.SUB}\end{tabular}}
 & macro-$\mathcal{F}$ & $\uparrow$  & \underline{0.620} & 0.575 & 0.591 & \textbf{0.636}~~~~~{$_{(+0.016)}$} 

\\
\midrule
 &$\text{Sum}~\mathcal{C}\mathcal{L}\mathcal{S}$& $\uparrow$ &\underline{309.02} & 298.28 & 307.82 & ~~~~~~~~\textbf{318.14}  \\
 \midrule
 & Bleu-4 & $\uparrow$ &3.30 & - &\underline{4.50} & \textbf{5.70}~~~~~~~$_{(+1.2)}$  \\
 & METEOR  & $\uparrow$  & 9.80 & - & \underline{10.9} & \textbf{12.8}~~~~~~~$_{(+1.9)}$ \\
\multirow{-4}{*}{\begin{tabular}[c]{@{}l@{}}\textsc{4.FC}\end{tabular}} 
& ROUGE-L  & $\uparrow$  & 29.7 &-& \underline{30.1} & \textbf{32.9}~~~~~~~$_{(+2.8)}$ \\
& CIDEr  & $\uparrow$  & 30.1 &- & \underline{30.7} & \textbf{32.6}~~~~~~~$_{(+1.9)}$ \\
\midrule
 &$\text{Sum}~\mathcal{C}\mathcal{A}\mathcal{P}$& $\uparrow$ &72.9 & -  & \underline{76.2} & ~~~~~~~~~\textbf{84.0}  \\
\bottomrule
\end{tabular}
}
\vspace{-10pt}
\end{table}


\begin{table*}[t!]
\center
\renewcommand{\tabcolsep}{2.5pt}
\renewcommand{\arraystretch}{1.3}
\scriptsize
\caption{\small \textbf{Ablation studies of 3 vital pre-training factors.} See \secref{sec:abaltion} for details.
}\label{tab:abalation}
\vspace{-10pt}
\resizebox{.98\textwidth}{!}
{
\begin{tabular}{lr||c|c|c|c|c|ccccccl}
\hline
\multicolumn{2}{c|}{\multirow{2}{*}{\textbf{\textsf{Metrics}}}} &
  \multicolumn{3}{c|}{\textbf{\textsf{KPG}}} &
  \multicolumn{2}{c|}{\textbf{\textsf{AGM}}} &
  \multicolumn{7}{c}{\textbf{\textsf{AKPM}}} \\ 
  \cline{3-14} 
\multicolumn{2}{c|}{} &
  Scale-fixed &
  Kaleido. &
  Kaleido.+SOD &
  Random &
  AGM &
  \multicolumn{1}{c|}{B} &
  \multicolumn{1}{c|}{B+I} &
  \multicolumn{1}{c|}{B+I$\sim$II} &
  \multicolumn{1}{c|}{B+I$\sim$III} &
  \multicolumn{1}{c|}{B+I$\sim$IV} &
  \multicolumn{1}{c|}{B+I$\sim$V} &
  \multicolumn{1}{c}{B+V} \\ 
  \hline
1. Rank@1&
   $\uparrow$ &
 24.71  & 
26.73(+8.2\%)   & 
 27.99(+13.3\%) &
  26.55 &
  27.99(+5.4\%) &
  \multicolumn{1}{c|}{25.37} &
  \multicolumn{1}{c|}{25.07(-1.2\%)} &
  \multicolumn{1}{c|}{26.03(+2.6\%)} &
  \multicolumn{1}{c|}{26.88(+6.0\%)} &
  \multicolumn{1}{c|}{26.20(+3.3\%)} &
  \multicolumn{1}{c|}{27.99(+10.3\%)} &
  24.62(-2.9\%) \\
  1. Rank@5&
   $\uparrow$ &
 50.05  & 
54.55(+9.0\%)   & 
 60.09(+20.1\%) &
  55.13 &
  60.09(+8.9\%) &
  \multicolumn{1}{c|}{54.97} &
  \multicolumn{1}{c|}{55.14(+0.3\%)} &
  \multicolumn{1}{c|}{56.31(+2.4\%)} &
  \multicolumn{1}{c|}{58.34(+6.1\%)} &
  \multicolumn{1}{c|}{59.13(+7.6\%)} &
  \multicolumn{1}{c|}{60.09(+9.3\%)} &
  53.78(-2.2\%) \\
  1. Rank@10&
   $\uparrow$ &
 58.93 & 
65.44(+11.0\%)   & 
 68.37(+16.0\%) &
  64.92 &
  68.37(+5.3\%) &
  \multicolumn{1}{c|}{62.13} &
  \multicolumn{1}{c|}{62.90(+1.2\%)} &
  \multicolumn{1}{c|}{63.37(+2.0\%)} &
  \multicolumn{1}{c|}{67.79(+9.1\%)} &
  \multicolumn{1}{c|}{67.99(+9.4\%)} &
  \multicolumn{1}{c|}{68.37(+10.0\%)} &
  60.88(-2.0)\% \\
  
 \hdashline
  2. Rank@1&
   $\uparrow$ &
 30.17  & 
32.19(+6.7\%)   & 
 33.88(+12.0\%) &
  32.14 &
  33.88(+5.4\%) &
  \multicolumn{1}{c|}{31.09} &
  \multicolumn{1}{c|}{30.98(-0.4\%)} &
  \multicolumn{1}{c|}{32.22(+3.6\%)} &
  \multicolumn{1}{c|}{33.17(+6.7\%)} &
  \multicolumn{1}{c|}{33.80(+8.7\%)} &
  \multicolumn{1}{c|}{33.88(+9.0\%)} &
  30.77(-1.0\%) \\
 2. Rank@5&
   $\uparrow$ &
 52.29  & 
58.40(+11.7\%)   & 
 60.60(+15.9\%) &
  56.99 &
  60.60(+6.3\%) &
  \multicolumn{1}{c|}{57.35} &
  \multicolumn{1}{c|}{57.44(+0.2\%)} &
  \multicolumn{1}{c|}{58.73(+2.4\%)} &
  \multicolumn{1}{c|}{58.55(+2.1\%)} &
  \multicolumn{1}{c|}{60.57(+5.6\%)} &
  \multicolumn{1}{c|}{60.60(+5.7\%)} &
  55.95(-2.4\%) \\
 2. Rank@10&
   $\uparrow$ &
 60.82 & 
66.49(+9.3\%)   & 
 68.59(+12.8\%) &
  63.77 &
  68.59(+7.6\%) &
  \multicolumn{1}{c|}{64.79} &
  \multicolumn{1}{c|}{65.65(+1.3\%)} &
  \multicolumn{1}{c|}{64.16(-1.0\%)} &
  \multicolumn{1}{c|}{67.92(+4.8\%)} &
  \multicolumn{1}{c|}{68.41(+5.6\%)} &
  \multicolumn{1}{c|}{68.09(+5.1\%)} &
 61.70(-4.8\%) \\
   \textbf{$\text{Sum}~\mathcal{R}$}&
   $\uparrow$ &
 \textbf{276.97} & 
\textbf{303.80(+9.7\%)}   & 
 \textbf{319.52(+16.2\%) }&
  \textbf{299.50 }&
  \textbf{319.52(+6.7\%)} &
  
  \multicolumn{1}{c|}{\textbf{295.70}} &
  \multicolumn{1}{c|}{\textbf{297.18(+0.5\%)}} &
  \multicolumn{1}{c|}{\textbf{300.82(+1.7\%)}} &
  \multicolumn{1}{c|}{\textbf{312.65(+5.7\%)}} &
  \multicolumn{1}{c|}{\textbf{316.10(+6.9\%)}} &
  \multicolumn{1}{c|}{\textbf{319.02(+7.9\%)}} &
  \textbf{287.70(-2.7\%)} \\
 \hline
3. ACC &
  $\uparrow$ &
 93.44\%  &
 93.45\%(+0.0\%)  &
 95.07\%(+1.7\%)  &
  92.71\% &
  95.07\%(+2.5\%) &
  \multicolumn{1}{c|}{90.94\%} &
  \multicolumn{1}{c|}{90.82\%(-0.1\%)} &
  \multicolumn{1}{c|}{91.40\%(+0.5\%)} &
  \multicolumn{1}{c|}{93.91\%(+3.3\%)} &
  \multicolumn{1}{c|}{94.05\%(+3.4\%)} &
  \multicolumn{1}{c|}{95.07\%(+4.5\%)} &
  88.87(-2.3\%) \\
  3. macro-$\mathcal{F}$ &
  $\uparrow$ &
 0.701  &
 0.705(+0.6\%)  &
 0.714(+1.9\%)  &
  0.711 &
  0.714(+0.4\%) &
  \multicolumn{1}{c|}{0.690} &
  \multicolumn{1}{c|}{0.692(+0.3\%)} &
  \multicolumn{1}{c|}{0.721(+4.5\%)} &
  \multicolumn{1}{c|}{0.713(+3.3\%)} &
  \multicolumn{1}{c|}{0.710(+2.9\%)} &
  \multicolumn{1}{c|}{0.714(+3.5\%)} &
  0.701(+1.4\%) \\
  \hdashline
 4. ACC &
  $\uparrow$ &
 86.89\%  &
87.61\%(+0.8\%)  &
 88.07\%(+1.4\%)  &
  87.20\% &
  88.07(+1.0\%) &
  \multicolumn{1}{c|}{81.66\%} &
  \multicolumn{1}{c|}{81.25\%(-0.5\%)} &
  \multicolumn{1}{c|}{84.44\%(+3.4\%)} &
  \multicolumn{1}{c|}{86.49\%(+5.9\%)} &
  \multicolumn{1}{c|}{88.53\%(+8.4\%)} &
  \multicolumn{1}{c|}{88.07\%(+7.9\%)} &
  81.64(+0.0\%) \\
 4. macro-$\mathcal{F}$ &
  $\uparrow$ &
 0.630  &
 0.634(+0.6\%)  &
 0.636(+1.0\%)  &
  0.633 &
  0.636(+0.5\%) &
  \multicolumn{1}{c|}{0.558} &
  \multicolumn{1}{c|}{0.575(+3.0\%)} &
  \multicolumn{1}{c|}{0.596(+6.8\%)} &
  \multicolumn{1}{c|}{0.636(+14.0\%)} &
  \multicolumn{1}{c|}{0.633(+13.4\%)} &
  \multicolumn{1}{c|}{0.636(+14.0\%)} &
  0.596(+8.4\%) \\
  \textbf{$\text{Sum}~\mathcal{C}\mathcal{L}\mathcal{S}$} &
  $\uparrow$ &
 \textbf{313.43}  &
 \textbf{314.96(+0.5\%)}  &
 \textbf{318.14(+1.5\%)}  &
  \textbf{314.31} &
  \textbf{318.14(+1.2\%)} &
  \multicolumn{1}{c|}{\textbf{297.40}} &
  \multicolumn{1}{c|}{\textbf{298.77(+0.4\%)}} &
  \multicolumn{1}{c|}{\textbf{307.54(+3.4\%)}} &
  \multicolumn{1}{c|}{\textbf{315.30(+6.0\%)}} &
  \multicolumn{1}{c|}{\textbf{316.88(+6.5\%)}} &
  \multicolumn{1}{c|}{\textbf{318.14(+7.0\%)}} &
  \textbf{300.21(+0.9\%)} \\
  \hline
 5. Bleu-4   &  $\uparrow$ &
 4.9  &
 5.2(+6.1\%)  &
 5.7(+16.3\%) &
 5.3  &
 5.7(+7.5\%)  &
  \multicolumn{1}{c|}{4.9} &
  \multicolumn{1}{c|}{5.2(+6.1\%)} &
  \multicolumn{1}{c|}{5.2(+6.1\%)} &
  \multicolumn{1}{c|}{5.1(+4.1\%)} &
  \multicolumn{1}{c|}{5.6(+14.3\%)} &
  \multicolumn{1}{c|}{5.7(+16.3\%)} &
  5.3(+8.2\%)\\
5. METEOR&  $\uparrow$ &
 11.0  &
 11.7(+6.4\%)  &
 12.8(+16.4\%) &
 11.3  &
 12.8(+13.3\%)  &
  \multicolumn{1}{c|}{11.6} &
  \multicolumn{1}{c|}{11.6(+0.0\%)} &
  \multicolumn{1}{c|}{11.8(+1.7\%)} &
  \multicolumn{1}{c|}{12.6(+8.6\%)} &
  \multicolumn{1}{c|}{12.8(+10.3\%)} &
  \multicolumn{1}{c|}{12.8(+10.3\%)} &
  11.4(-1.7\%)\\
   5. ROUGE-L  &  $\uparrow$ &
 29.8  &
 31.5(+5.7\%)  &
 32.9(+10.4\%) &
 30.3  &
 32.9(+8.6\%)  &
  \multicolumn{1}{c|}{30.4} &
  \multicolumn{1}{c|}{30.7(+1.0\%)} &
  \multicolumn{1}{c|}{30.8(+1.3\%)} &
  \multicolumn{1}{c|}{31.9(+4.9\%)} &
  \multicolumn{1}{c|}{32.7(+7.6\%)} &
  \multicolumn{1}{c|}{32.9(+8.2\%)} &
  30.6(+0.7\%)\\
   5. CIDEr  &  $\uparrow$ &
 30.9  &
 31.3(+1.3\%)  &
 32.6(+5.5\%) &
 31.7  &
 32.6(+2.8\%)  &
  \multicolumn{1}{c|}{31.0} &
  \multicolumn{1}{c|}{31.5(+1.6\%)} &
  \multicolumn{1}{c|}{31.4(+1.3\%)} &
  \multicolumn{1}{c|}{32.0(+3.2\%)} &
  \multicolumn{1}{c|}{32.3(+4.2\%)} &
  \multicolumn{1}{c|}{32.6(+5.2\%)} &
  31.3(+1.0\%)\\
  \textbf{$\text{Sum}~\mathcal{C}\mathcal{A}\mathcal{P}$} &  $\uparrow$ &
 \textbf{76.6}  &
 \textbf{79.7(+4.0\%)}  &
 \textbf{84.0(+9.7\%)} &
 \textbf{78.6 } &
 \textbf{84.0(+6.9\%)}  &
  \multicolumn{1}{c|}{\textbf{77.9}} &
  \multicolumn{1}{c|}{\textbf{79.0(+1.4\%)}} &
  \multicolumn{1}{c|}{\textbf{79.2(+1.7\%)}} &
  \multicolumn{1}{c|}{\textbf{81.6(+4.7\%)}} &
  \multicolumn{1}{c|}{\textbf{83.4(+7.1\%)}} &
  \multicolumn{1}{c|}{\textbf{84.0(+7.8\%)}} &
  \textbf{78.6(+0.9\%)}\\  
  
  \hline
\end{tabular}
}\label{tab:AblationStudy1}
\end{table*}

\begin{figure*}[t!]
	\centering
	\includegraphics[width=1\textwidth]{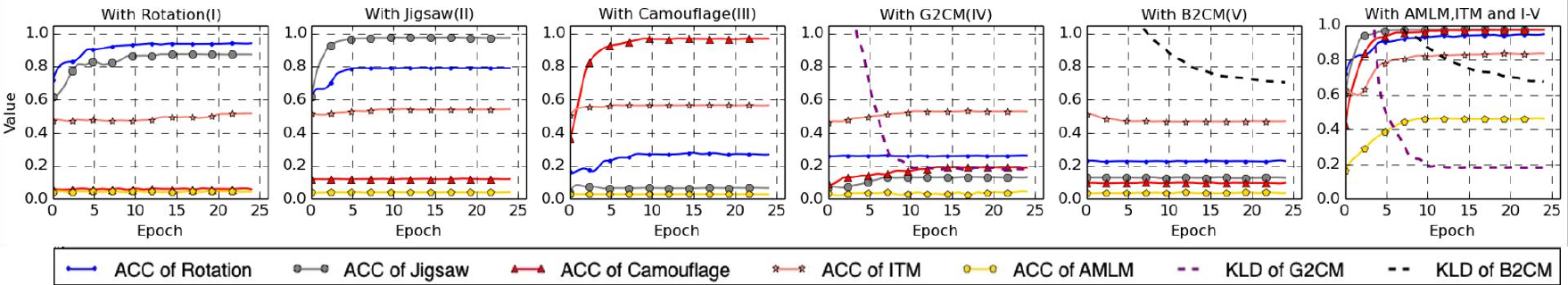} 
	\vspace{-15pt}
	\caption{\small \textbf{Single task analysis on our \ourmodel.} ACC = Accuracy, KLD = Kullback-Leibler Divergence.
	Refer to \secref{sec:abaltion} for details.
	}\label{fig:curve}
\end{figure*}

\subsection{Ablation Study}\label{sec:abaltion} 
Three main factors may influence the performance of \ourmodel, including Input-level: Kaleido Patch Generator (KPG); Embedding-level: 
Alignment Guided Masking (AGM); and Task-level: Aligned Kaleido Patch Modeling (AKPM). We thus perform three main ablation studies to further analyze these components/strategies of our model. The results are shown in Tab.~\ref{tab:AblationStudy1} and Fig.~\ref{fig:curve}.

\myPara{KPG.~}
Three attempts we have tried to generate our kaleido patches. \textbf{Scheme-1:} Similar to~\cite{gao2020fashionbert, vit2021ICLR}, the first attempt is to split the fashion images with a fixed-scale setting. 
Training with such patches, we obtain 276.97 Sum~$\mathcal{R}$, 313.43 Sum~$\mathcal{C}\mathcal{L}\mathcal{S}$ and 76.6 Sum~$\mathcal{C}\mathcal{A}\mathcal{P}$ scores.
\textbf{Scheme-2:} We carry out a kaleido (patch-invariant) scheme to generate patches and achieves +9.7\%, +0.5\% and +4.0\% relative improvement on each metric. Compared with scheme-1, scheme-2 is capable of capturing fine-grained representation better.
\textbf{Scheme-3:} We further introduce the salient object detection (SOD) algorithm~\cite{zhao2019egnet} to avoid a huge number of patches with blank regions (tabula rasa). We observed 16.2\%,  +1.5\% and +9.7\% relative improvement compared with Scheme-1.

\myPara{AGM.~} 
The majority of existing masking methods independently mask visual and text features with a pre-set probability.
Such kind masking methods are usually named as random masking. 
In this experiment, we compare AGM to random masking (Random). 
Not surprisingly, AGM obtains +6.7\%, +1.2\%, 6.9\% improvements.
Compared to random masking, AGM generates more semantic related masking, 
which benefits our~\ourmodel~to better understand multi-modality information.

\myPara{AKPM.~} 
To verify the efficiency of the proposed AKPM, we conduct 7 ablation studies (see \figref{fig:curve}). The baseline (B) merely consists of the basic ITM and AMLM. Then we add five sub-task to pre-train the model step by step. For example, ``B + I $\sim$ IV'' equals to ``B + I + II + III + IV''. 
Note that existing models~\cite{gao2020fashionbert}~usually use the combination of ``ITM + AMLM + B2CM'' (B + V) as the pre-training supervision. As shown in \tabref{tab:abalation}, the improvement is limited (+0.9\%) in terms of  $\text{Sum}~\mathcal{C}\mathcal{L}\mathcal{S}$ score, even cause negative effect (-2.7\%) in $\text{Sum}\mathcal{R}$ metric. Interestingly, we naively replace V (B2CM) with I (RR) will obtains improvement on all downstream tasks (+0.5\%, +0.4\% and +1.4\%, respectively). Gradually, we observed that the performance continue improve when we adding the corresponding sub-tasks sequentially.
Meanwhile the negative affect of V has been alleviated, we argue that V plays the true value when \ourmodel~has learned the comprehensive representations of image embeddings from I$\sim$IV.

%



\section{Conclusion}
We presented a novel pre-trained vision-language understanding 
architecture \textbf{\ourmodel} for fashion-based task. It consists of a kaleido patches generator, attention-based alignment generator, and alignment guided masking strategy. 
These components are easy to implement and cooperate closely to learn both intra-modal and inter-modal image-text feature embeddings. 
The designed  \ourmodel~is much more efficient than existing models, attains the new SOTA performance, and largely boosts the accuracy of many downstream tasks such as Image-Text Retrieval, Category Recognition, and Fashion Captioning. 


{\small
\bibliographystyle{ieee_fullname}
\bibliography{kaleidobert}
}


\end{document}